\documentclass[conference]{IEEEtran}
\IEEEoverridecommandlockouts
\usepackage{cite}
\usepackage{soul}
\usepackage{url}
\usepackage[hidelinks]{hyperref}
\usepackage[utf8]{inputenc}
\usepackage[small]{caption}
\usepackage{graphicx}
\usepackage{amsmath}
\usepackage{booktabs}
\usepackage{amsmath,amssymb,amsfonts}
\usepackage{algorithmic}
\usepackage{graphicx}
\usepackage{textcomp}
\usepackage{xcolor}
\usepackage{longtable}% for long tables
\usepackage{lineno,hyperref}
\usepackage{xcolor,stfloats}
\usepackage{algorithm}% http://ctan.org/pkg/algorithms
\usepackage{verbatim}
\usepackage{comment}
\usepackage{pifont}
\usepackage{times}
\usepackage{tikz,xcolor,hyperref}% Make Orcid icon
\definecolor{lime}{HTML}{A6CE39}
\DeclareRobustCommand{\orcidicon}{%
    \begin{tikzpicture}
    \draw[lime, fill=lime] (0,0) 
    circle [radius=0.16] 
    node[white] {{\fontfamily{qag}\selectfont \tiny ID}};    \draw[white, fill=white] (-0.0625,0.095) 
    circle [radius=0.007];    \end{tikzpicture}
    \hspace{-2mm}}
\foreach \x in {A, ..., Z}{%
    \expandafter\xdef\csname orcid\x\endcsname{\noexpand\href{https://orcid.org/\csname orcidauthor\x\endcsname}{\noexpand\orcidicon}}
    }
\def\BibTeX{{\rm B\kern-.05em{\sc i\kern-.025em b}\kern-.08em
    T\kern-.1667em\lower.7ex\hbox{E}\kern-.125emX}}

\begin{document}

\title{Author Name Disambiguation via Heterogeneous Network Embedding from Structural and Semantic Perspectives}
%: Improvement on The Second Prize of IJCAI2021-WhoIsWho Competition
%\\

\iffalse
\author{
\IEEEauthorblockN{Wenjin Xie\orcidA{}}
\IEEEauthorblockA{\textit{College of Computer and Information Science}\\
\textit{Southwest University}\\
Chongqing, China \\
xiewenjin@email.swu.edu.cn}
\\
\IEEEauthorblockN{Xiaomeng Wang\orcidB{}}
\IEEEauthorblockA{\textit{College of Computer and Information Science}\\
\textit{Southwest University}\\
Chongqing, China \\
wxm1706@swu.edu.cn\\}
\and
\IEEEauthorblockN{Siyuan Liu\orcidD{}}
\IEEEauthorblockA{\textit{College of Computer and Information Science}\\
\textit{Southwest University}\\
Chongqing, China \\
liusiyuan@email.swu.edu.cn\\}
\\
\IEEEauthorblockN{Tao Jia\IEEEauthorrefmark{1}\orcidC{}}
\IEEEauthorblockA{\textit{College of Computer and Information Science}\\
\textit{Southwest University}\\
Chongqing, China \\
tjia@swu.edu.cn\\}
%\IEEEauthorrefmark{1}Tao Jia is the corresponding author.
}
\fi

\author{
\IEEEauthorblockN{Wenjin Xie\orcidA{}, Siyuan Liu\orcidD{}, Xiaomeng Wang\orcidB{}, Tao Jia\orcidC{}\IEEEauthorrefmark{2}}
\IEEEauthorblockA{College of Computer and Information Science\\
Southwest University\\
Chongqing, China\\
xiewenjin,liusiyuan@email.swu.edu.cn; wxm1706,tjia@swu.edu.cn\\}
}

\maketitle

\begin{abstract}

Name ambiguity is common in academic digital libraries, such as multiple authors having the same name. This creates challenges for academic data management and analysis, thus name disambiguation becomes necessary. The procedure of name disambiguation is to divide publications with the same name into different groups, each group belonging to a unique author. A large amount of attribute information in publications makes traditional methods fall into the quagmire of feature selection. These methods always select attributes artificially and equally, which usually causes a negative impact on accuracy. The proposed method is mainly based on representation learning for heterogeneous networks and clustering and exploits the self-attention technology to solve the problem. The presentation of publications is a synthesis of structural and semantic representations. The structural representation is obtained by meta-path-based sampling and a skip-gram-based embedding method, and meta-path level attention is introduced to automatically learn the weight of each feature. The semantic representation is generated using NLP tools. Our proposal performs better in terms of name disambiguation accuracy compared with baselines and the ablation experiments demonstrate the improvement by feature selection and the meta-path level attention in our method. The experimental results show the superiority of our new method for capturing the most attributes from publications and reducing the impact of redundant information.

\end{abstract}


\section{Introduction}

Name ambiguity is a prevalent problem in scholarly publications due to the unprecedented growth of digital libraries and the number of researchers. An author is identified by their name in the absence of a unique identifier. This presents a problem called author name ambiguation caused by the same name spellings. The existence of ambiguities deteriorates the performance of document retrieval and web search, which can also mislead to an improper assessment of the authors. The author name disambiguation aims to assign different publications with identical author name to some specific academic researchers, who are their actual owners. Further, it can affect various bibliometric analysis tasks like co-authorship link prediction \cite{wang2014link}, collaboration network analysis \cite{viana2013time}, and so on. 

The author name disambiguation task is considered as a clustering problem in the academic network in recently mainstream researches \cite{zhang2018name,xu2018network,qiao2019unsupervised}. Its goal is to assign publications with the same author name to different clusters, so that all publications of a cluster belong to one specific author. The publications are clustered by their similarity, based on the fact that all of an author's publications should be of similarity to some degree. To lay a good foundation for clustering, recent works have focused on the development of low-dimensional representations to create feature spaces to reduce the heterogeneity of entity information \cite{zhang2017name,zhang2018name,ictaizukov2017neural,qiao2019unsupervised,pooja2022exploiting}. Since these approaches rely on predefined relationships to build connections between publications, there seems to be a contradiction between comprehensiveness and efficiency. If considering as many relationships between publications as possible, some redundant information will be imported to intervene in the accuracy and increase the cost. On the other hand, artificially selecting a part of relationships may lead to the loss of valuable features and is highly dependent on the empirical selection. Self-attention happens to address this problem. It can capture the different importance of different features and assign them the corresponding weights when aggregating them \cite{attention,gat,han}. In name disambiguation task, the importance of each relation can be learned automatically using an attention algorithm, so that the more critical relational features could get more attention.

The semantic features also matter as the publications of one author are always similar in some text features, like research topics and writing styles \cite{pooja2021exploiting}. Compared to the works only considering the structural features of publications, the works using semantic features like the text information of title, abstract and keywords of publications show more accuracy in name disambiguation \cite{xu2018network,qiao2019unsupervised,pooja2022exploiting}.

In this paper, we introduce our author name disambiguation method which takes both structural and semantic perspectives into consideration to embed publications to vectors and cluster them based on the embeddings. From a structural perspective, our method exploits a weighted meta-path walk to sample context publications, capturing the heterogeneity of each meta-path. We propose the meta-path level attention to obtain the weights of various meta-paths and jointly learn the overall structural publication representations. From a semantic perspective, our new method uses Doc2Vec technology to get representations, which is proven to deal well with text embedding tasks of publications \cite{doc2vec}. Both publication representations are combined and clustered by an adaptive clustering method. The performance of our method is verified to be significantly better than several state-of-the-art methods.

\section{Related Works}

The works addressing author name disambiguation can be distinguished on the method-level of machine learning approaches and non-machine-learning approaches\cite{hussain2017survey}.
Machine learning approaches can implicitly discover the same statistical characteristics of publications and their attributes which are characterized by their training data. The supervised methods require external labeling of training data regarding whether the papers in question are from the same author. With this information, the methods learn how the attributes refer to the same or different authors \cite{kim2019hybrid,kim2019generating,rehs2021supervised}. Unsupervised methods try to find latent patterns in the data by themselves rather than using labels but are therefore often short of accuracy and efficiency \cite{wu2014unsupervised,qiao2019unsupervised,d2020collecting}.
The non-machine-learning-based approaches are split into heuristic-based and graph-based methods. Heuristic approaches use paper attributes to construct simple rules with which authors and their papers can be distinguished \cite{de2011incremental,tang2011unified}.
Graph-based methods use publications and their attributes as node and edge representations to detect author communities in a graph. As a result, the publications with the same author name but different author entities can be separated into various connected graph-communities \cite{on2012scalable,shin2014author}. With the development of network embedding technology on both homogeneous and heterogeneous graphs \cite{deepwalk,tang2015line,metapath2vec,zhan2022coarsas2hvec}, the graph-based method gains a massive promotion. The network consists of node attributes and connections can be represented by vectors thus the relationship between nodes can be captured by the similarity of their embeddings \cite{zhang2018name,qiao2019unsupervised,pooja2021exploiting,pooja2022exploiting}.

Attention mechanisms, like self-attention \cite{attention} and soft-attention \cite{bahdanau2014neural}, have become one of the most influential technologies in deep learning. In network embedding works, Graph Attention Networks is proposed to learn the importance between nodes and their neighbors and fuse the neighbors to perform node classification \cite{gat}. Heterogeneous Graph Attention Networks introduces node-level attention to aggregate the neighbors' embeddings to the node and semantic-level attention to aggregate embeddings of different semantic meta-paths \cite{han}.
Self-attention has been introduced for the entity recognition and name disambiguation task. \cite{ictaizukov2017neural} uses self-attention to learn the corresponding weights of different features of an entity to improve the entity recognition performance. \cite{pooja2022exploiting} proposes to exploit self-attention to automatically learn the corresponding weights of different neighbors of a node in the publication network and also distinguish the importance of co-author and meta-content relationships. It is verified in these works that the attention mechanism can improve the name disambiguation to some degree, but these works use the attention mostly on the node level. It remains to investigate to use the attention technology to learn the importance of different relational attributes.

\section{Preliminary}
\label{section:P}
The author name disambiguation task derives from the academic data, including authors, papers, venues and so on, which could construct an academic heterogeneous network. In this section, we introduce some definitions about the name disambiguation task and the publication heterogeneous network.

\textbf{Definition 1: Author name disambiguation.} For a given ambiguous author name $a$, $\mathcal{P}^{a}=\left\{P_{1}^{a}, P_{2}^{a}, \ldots, P_{N}^{a}\right\}$ are the set consisting of all the publications whose authors include $a$. The author name disambiguation task aims to divide the publications in $\mathcal{P}^{a}$ into $S$ different clusters $\left\{C_{1}^{a}, C_{2}^{a}, \ldots, C_{S}^{a}\right\}$ so that the publications in a cluster are written by the same author and the ones in different clusters belong to different authors.

\begin{figure}[!htb]
	\centering
	\includegraphics[width=0.8\linewidth]{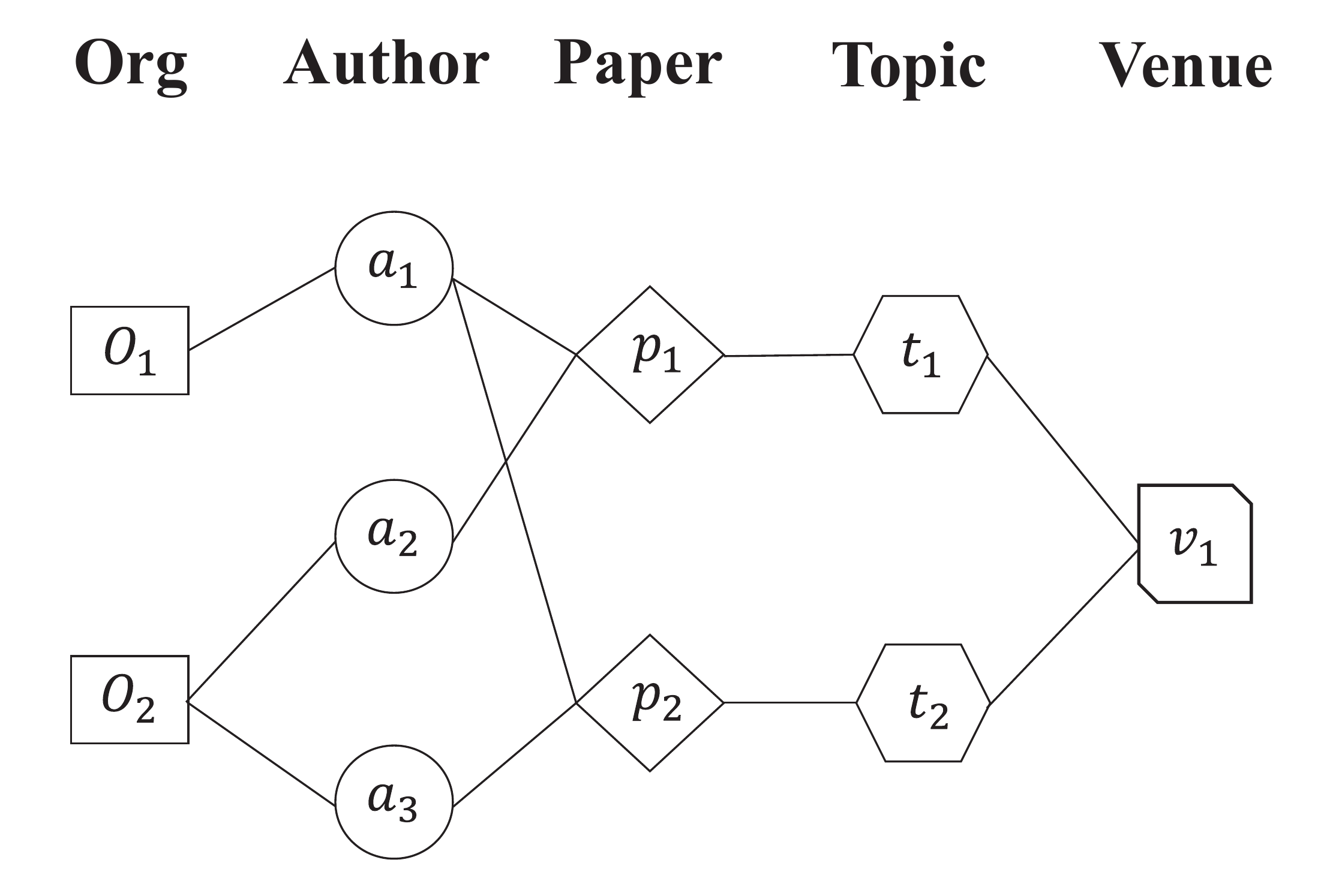}
	\caption{An example of academic heterogeneous network.
\label{fig:pub-hn}}
\end{figure}

\textbf{Definition 2: Publications heterogeneous network.} All the items in the academic data, including authors, papers, venues and so on, could construct an academic heterogeneous network like Fig.\ref{fig:pub-hn}. The publication heterogeneous network is extracted from the academic network, whose nodes are all publications and edges are multiple relations among them. For every ambiguous author name $a$, its publications heterogeneous network can be expressed as $G^{a}=(V, E, \mathcal{R})$, where $\mathcal{R}$ stands for co-occurrence relationships between two publications. For instance, if two publications get the same author name, we define there is a co-author relationship between these two publications. 
%Analogously, there are also relationships like co-venue (two documents are published on the same venue), co-year (two documents are published in the same year) and so on. 

\textbf{Definition 3: Weighted Meta-path walk.} The meta-path walk is an artificial sampling strategy of nodes in heterogeneous networks \cite{metapath2vec, pathsim}. Different from the random walk strategy which simply samples nodes along the edges of the network randomly, the meta-path walk specifies the type of the node sampled at each time, so the edge of each walk is not completely random. Symmetric meta-paths are widely used for they can capture the co-occurrence relations between nodes. For example, giving the meta-path Paper-Author-Paper, the sampling walk must jump to the node whose type is author when starts from the paper node. 
%In Fig.\ref{fig:pub-hn}, if the walk starts from node $p_1$, the sampling node sequence must be $p_1-a_1-p_2$. Obviously, this meta-path demonstrates the two papers belong to the same author. The weights of the edges are also considered, which depend on the degree of co-occurrence relationship. If there are three co-authors between papers $p_1$ and $p_2$ while one co-author between $p_1$ and $p_3$, the probability of sampling $p_1-p_2$ is 3 times the probability of sampling $p_1-p_3$, when we walk starting from $p_1$ following the meta-path Paper-Author-Paper.

We summarize the list of important notations used in this paper in Tab.\ref{tab:notations}.

\begin{table}[htbp]
  \centering
  \caption{List of notations}
    \begin{tabular}{c|l}
    \hline
    Symbol & \multicolumn{1}{c}{Definition} \\
    \hline
    $a$     & Ambiguous author name \\
    $\mathcal{P}^{a}$    & Publication set of ambiguous name $a$ \\
    $C^a$     & Cluster set predicted for $a$  \\
    $G^a$     & Publication heterogeneous network of $a$ \\
    $p_i$     & Publication(node) $i$ in $G^a$ \\
    $z_i$     & Embedding of $p_i$ \\
    $z_{i}^{0}$     & Initial structural vector for $p_i$ \\
    $\Phi_n$     & Meta-path $n$ \\
    $Z_{\Phi_n}$     & Overall node embedding for meta-path $n$ \\
    $z_{\Phi_n}^{i}$     & Node $i$'s embedding for $\Phi_n$ \\
    $\alpha_{\Phi_n}$     & Importance weight of $\Phi_n$ \\
    $w_{\Phi_n}$     & Attention coefficient of $\Phi_n$ \\
    $Z$     & Final structural embedding \\
    $A$     & Adjacent matrix corresponding to homogeneous network \\
    $\hat{A}$     & Regenerated adjacent matrix of $A$ \\   
    $M^{st}$     & Adjacent structural similarity matrix \\
    $M^{se}$     & Adjacent semantic similarity matrix \\
    $M$     & Overall adjacent similarity matrix \\
    \hline
    \end{tabular}%
  \label{tab:notations}%
\end{table}%

\section{Methodology}

In this section, we introduce our method for the author name disambiguation task. The method captures both the structural feature and semantic feature of a publication and generates a representation of it. The weighted sum of two embeddings is obtained and the node embeddings are clustered using the adaptive clustering method.

\subsection{Publication Representation based on Structural Feature}

The most commonly used method to capture the structural features of nodes is network embedding based on the network walk strategy, which is sampling the nodes along the edges of the network to learn the correlation between two neighbor nodes. 
As there are various relations between publications in the academic heterogeneous network, the meta-path-based network embedding method are considered in our work. The meta-path walk, as introduced in Section \ref{section:P}, is able to capture the features of heterogeneous nodes and thus is widely used in heterogeneous network representation learning \cite{metapath2vec,pathsim,han,ictaisong2018meta,ictaizhang2020meta}.% 

As the author name disambiguation task is focused on the representations of publications, the mate-paths we exploit here are about the relationships between papers. To be specific, we select Paper-Author-Paper(PAP), Paper-Organization-Paper(POP), Paper-Venue-Paper(PVP), Paper-Year-Paper(PYP), and Paper-Word-Paper(PWP). Based on these meta-paths, the representations of publications can be learned by a skip-gram model like what the former works did \cite{deepwalk,metapath2vec}. For the ambiguous publication network $G^{a}=(V, E, \mathcal{R})$ defined in Section \ref{section:P}, the node embeddings can be learned by maximizing the probability of any node $p_i$ having its neighbor $p_c$:

\begin{equation}
\arg \max _{\theta} \sum_{p_{i} \in V} \sum_{r \in \mathcal{R}} \sum_{p_{c} \in N_{r}\left(p_{i}\right)}w_{p_{c}} \log p\left(p_{c} \mid p_{i}; \theta\right), \label{eq:skip-gram}
\end{equation}
where $N_{r}\left(p_{i}\right)$ is the neighbor set of $p_i$, and $w_{p_{c}}$ is the weight of the edge $(p_i,p_c)$ to make sure the neighbors who are more intimate with $p_i$ can get a higher probability output. $\log p\left(p_{c} \mid p_{i}; \theta\right)$ is commonly defined as a softmax function \cite{mikolov2013distributed,bengio2013representation,goldberg2014word2vec},
that is: 
\begin{equation}
p\left(p_{c} \mid p_{i} ; \theta\right)=\frac{e^{z_{i} \cdot z_{c}}}{\sum_{u \in V} e^{z_{i} \cdot z_{c}}}, \label{eq:softmax}
\end{equation}
where $z_i$ is the node representation of $p_i$ generated from its initial vector $z_i^{0}$. Here We use Word2Vec \cite{goldberg2014word2vec} to generate the initial vector of the publication $z_i^{0}$ based on the features corresponding to the meta-path. For example, given the meta-path POP, we input the organization name as a word into Word2Vec model to obtain $z_i^{0}$.
Then the most commonly used negative sampling method is leveraged to achieve efficient optimization \cite{mikolov2013distributed,metapath2vec,qiao2019unsupervised}, and the probability in Eq.\ref{eq:skip-gram} can be updated as follows:

\begin{equation}
\begin{split}
\log p\left(p_{c} \mid p_{i}; \theta\right) \approx \log \sigma\left(z_{i} \cdot z_{c}\right)\\ +\sum_{t=1}^{T} \mathbb{E}_{p^{t} \sim P(u)}\left[\log \sigma\left(-z_{i} \cdot z_{t}\right)\right],\label{neg}
\end{split}
\end{equation}
where $\sigma(x)=\frac{1}{1+e^{-x}}$ and $P(u)$ is the pre-defined distribution from which a negative sampling node $p_t$ is drew for $T$ times. $z_t$ is the embedding of $p_t$.
For each meta-path $\Phi$, the node embedding can be learned by the method above. Thus given the meta-path set $\{\Phi_{1},\Phi_{2},...,\Phi_{m}\}$($m=4$ in this work), $m$ groups of structural specific node embeddings can be obtained, denoted as $\{Z_{\Phi_{1}},Z_{\Phi_{2}},...,Z_{\Phi_{m}}\}$. 

As different meta-paths have different effects on network embedding, each single node embedding $Z_{\Phi_{n}} (n\in [1,m])$ can only reflect the structural feature of nodes from one aspect. In order to learn a more comprehensive and accurate node embedding, we need to fuse multiple representations that obtained through different meta-paths. However, there remains a question that not all the relations matters equally when embedding. The co-author is a relatively stable feature in name disambiguation task \cite{qiao2019unsupervised,pooja2022exploiting}, while two publications having the similar topics does not mean they are written by the same author, since researchers sometimes change their research directions \cite{jia-nhb}. The fact presents the challenge of taking advantage of as many meta-paths as possible while eliminating the interference of redundant information brought by relations. To solve the challenge of meta-path selection and fusion in heterogeneous networks, we introduce a meta-path level attention to automatically learn the importance of different meta-paths. The weights of all meta-paths $(\alpha_{\Phi_{1}},\alpha_{\Phi_{2}},...,\alpha_{\Phi_{m}})$ can be calculated when the node embeddings corresponding to meta-paths are known:
\begin{equation}
(\alpha_{\Phi_{1}},\alpha_{\Phi_{2}},...,\alpha_{\Phi_{m}}) = att_{mp}(Z_{\Phi_{1}},Z_{\Phi_{2}},...,Z_{\Phi_{m}}),
\end{equation}
where $att_{mp}$ denotes the deep neural network using meta-path level attention. To derive the importance for each meta-path, we let the node $p_i$'s embedding of meta-path $\Phi_n$, denoted as $z_{\Phi_{n}}^{i}$, to pass through a single layer dense network with an activation function and obtain the output embedding for one node: $\tanh{\left(\mathbf{W} \cdot z_{\Phi_{n}}^{i}+\mathbf{b}\right)}$. Here $\mathbf{W}$ is the weighted matrix of the dense network, $\mathbf{b}$ is the bias vector. The attention coefficient $w_{\Phi_{n}}$ for meta-path $\Phi_{n}$ is then derived by averaging the representations of all nodes in the network as
\begin{equation}
w_{\Phi_{n}}=\frac{1}{|\mathcal{V}|} \sum_{i \in \mathcal{V}} \mathrm{q}^{\mathrm{T}} \cdot \tanh \left(\mathbf{W} \cdot z_{\Phi_{n}}^{i}+\mathbf{b}\right), \label{eq:mp att}
\end{equation}
where $\mathcal{V}$ is the node set in $\mathcal{G}$, $\mathrm{q}$ is the meta-path level attention vector. All the parameters above are shared for all meta-paths and the weight of meta-path $\Phi_n$ is derived by normalizing the importance of all meta-paths using softmax function:
\begin{equation}
\alpha_{\Phi_{n}}=\frac{\exp (w_{\Phi_{n}})}{\sum_{n=1}^{m} \exp (w_{\Phi_{n}})}. \label{eq:all att}
\end{equation}
With the weights of each meta-path learned, the overall embedding $Z$ can be derived by fuse these embedding corresponding to specific meta-paths:
\begin{equation}
Z= \sum_{n=1}^{m} \alpha_{\Phi_{n}} \cdot Z_{\Phi_{n}}. \label{eq:final st embedding}
\end{equation}

%the publication heterogeneous network is projected to the different embeddings based on different meta-paths, then the weight of each meta-path is jointly learned and the overall embeddings are fused via meta-path level attention.
\iffalse
The overall framework of encoding the embedding process is shown as Fig.\ref{fig:meta-path attention}. 
\begin{figure}[!htb]
	\centering
	\includegraphics[width=0.95\linewidth]{meta-path attention.pdf}
	\caption{The framework of the encoding process of publication representation learning based on structural features. $\Phi_{n}, n\in [1,m]$ are the different meta-paths. $Z_{\Phi_{n}}, n\in [1,m]$ are the corresponding embeddings generated by skip-gram for all meta-paths. $Z$ is the overall publication representation aggregated by the whole embeddings.
\label{fig:meta-path attention}}
\end{figure}
\fi

The decoder of the name disambiguation model aims to regenerate the original homogeneous network, for there is no influence of various relations between publications, same as that in \cite{pooja2022exploiting}. The model then can be trained by the linkage information between nodes in the homogeneous graph. For the loss calculation in the learning process, we use the expected reconstruction loss corresponding to the homogeneous graph. The expected reconstruction loss is represented by the negative log-likelihood of $\hat{A}$, where the expectation is taken with respect to the encoders distribution over the representations $\rm q(Z \mid X, A)$. The loss function, which attempt to make regenerated matrix $\hat{A}$ approach $A$, can be calculated as

\begin{equation}
\mathcal{L}=-\mathbb{E}_{q(Z \mid X, A)}[\log p(\hat{A} \mid Z)]. \label{loss}
\end{equation}

With the guide of labeled data, we can optimize the proposed model via back propagation and learn the final structural embeddings of publications. Based on the embeddings, the adjacent structural similarity matrix $M^{st}$ of publications is constructed, and the value $M^{st}_{i,j}$ means the structural similarity of publication $i$ and $j$, which is the cosine similarity of their embeddings.

\subsection{Publication Representation based on Semantic Feature}
The title, abstract and keywords are considered as the semantic features of a publication, thus we collect these three parts of all publications and learn their embeddings using the NLP method. In this work, we get the semantic publication representations using the Doc2Vec \cite{doc2vec} instead of Word2Vec \cite{mikolov2013distributed} which is used in our former method in the competition. Doc2Vec is proved to be more accurate than Word2Vec in calculating the semantic representation of publications \cite{doc2vec,qiao2019unsupervised}. Based on the semantic presentations, we construct the semantic adjacent matrix $M^{se}$ for all publications, where the value $M^{se}_{i,j}$ means the semantic similarity of publication $i$ and $j$, which is the cosine similarity of their semantic embeddings.

\subsection{Clustering Algorithm}
The publication similarity matrix $M$ can be finally obtained by combining the structural adjacent matrix $M^{st}$ and the semantic adjacent matrix $M^{se}$ with wights. We cluster the publications based on similarity matrix $M$ using hierarchical density-based spatial clustering method (HDBSCAN) and affinity propagation clustering method (AP) which does not need the prior knowledge of cluster numbers. The final result of clustering is determined adaptively based on the results of both two algorithms, proposed by \cite{xu2018network}.

\section{Experiments}
As our author name disambiguation method exploits the \textbf{M}eta-path level \textbf{A}ttention based embedding using \textbf{S}tructural and \textbf{S}emantic features, we name our method as MASS. MASS is compared with some baselines in experiments.

\subsection{Baselines and Experimental Data}
 To validate the performance of MASS we proposed, it is compared with several baseline methods, including \textbf{Zhang et al.}\cite{zhang2018name}, \textbf{Xu et al.}\cite{xu2018network}, \textbf{Qiao et al.}\cite{qiao2019unsupervised}, \textbf{Pooja et al.}\cite{pooja2022exploiting} and some embedding method like \textbf{DeepWalk}\cite{deepwalk}, \textbf{Metapath2Vec}\cite{metapath2vec} and \textbf{GraphSAGE}\cite{graphsage}. Two commonly used benchmarks are exploited in the experiment, i.e., the Aminer WhoIsWho data \cite{tang2008arnetminer,wang2011adana,tang2011unified} in the IJCAI2021 name disambiguation competition\footnote{https://www.biendata.xyz/competition/who-is-who2021/} and CiteSeeX data\footnote{http://clgiles.ist.psu.edu/data/}. The most commonly used metric Macro Pairwise F1 is explored for model evaluation, which is determined by whether every two publications predicted to the same author are indeed written by the same author.

\subsection{Experimental Results}
The experimental results of our MASS model and baselines on Aminer and CiteSeeX are illustrated in Tab.\ref{tab:sota}. Note that the parameters of all methods are fine-tuned to touch the best performances.

\begin{table}[htbp]
  \centering
  \caption{Performances of different methods on Aminer and CiteSeeX data set.}
    \begin{tabular}{ccc}
    \toprule
    Method & Aminer & CiteSeeX \\
    \midrule
    Zhang et al. & 0.866 & 0.538 \\
    Xu et al. & 0.848 & 0.646 \\
    Qiao et al. & 0.880 & 0.698 \\
    Pooja et al. & 0.879 & 0.702 \\
    \midrule
    DeepWalk & 0.835 & 0.429 \\
    Metapath2Vec & 0.875 & 0.547 \\
    GraphSAGE & 0.871 & 0.523 \\
    \midrule
    MASS  & \textbf{0.897} & \textbf{0.719} \\
    \bottomrule
    \end{tabular}%
  \label{tab:sota}%
\end{table}%

The comparison clearly shows that MASS performs better than baselines, demonstrating the superiority of MASS by leveraging both structural and semantic features and using meta-path level attention when embedding. The performance of Xu et al.\cite{xu2018network} verifies the importance of semantic features in name disambiguation task. MASS gets a higher F1 score than the method of Qiao et al.\cite{qiao2019unsupervised}, which processes the different relations equally, proving our method captures the different importance of different relations. While considering enough relationships, MASS does not allow redundant information to interfere with the results by using the attention technology. Compared to the network-embedding-based methods, the superiority of MASS to DeepWalk and GraghSAGE indicates that our heterogeneous network embedding method is more suitable when representing the publications in the academic network, and it also verifies the weighted meta-path guided walk brings a large benefit to skip-gram and random walk based network embedding method. Metapath2Vec does not import different relation weights when embedding, while MASS allocates adaptively weights to all meta-paths based on the self-attention mechanism, which also promotes performance.

\subsection{Ablation Analysis}
\begin{figure}[!htb]
	\centering
	\includegraphics[width=0.9\linewidth]{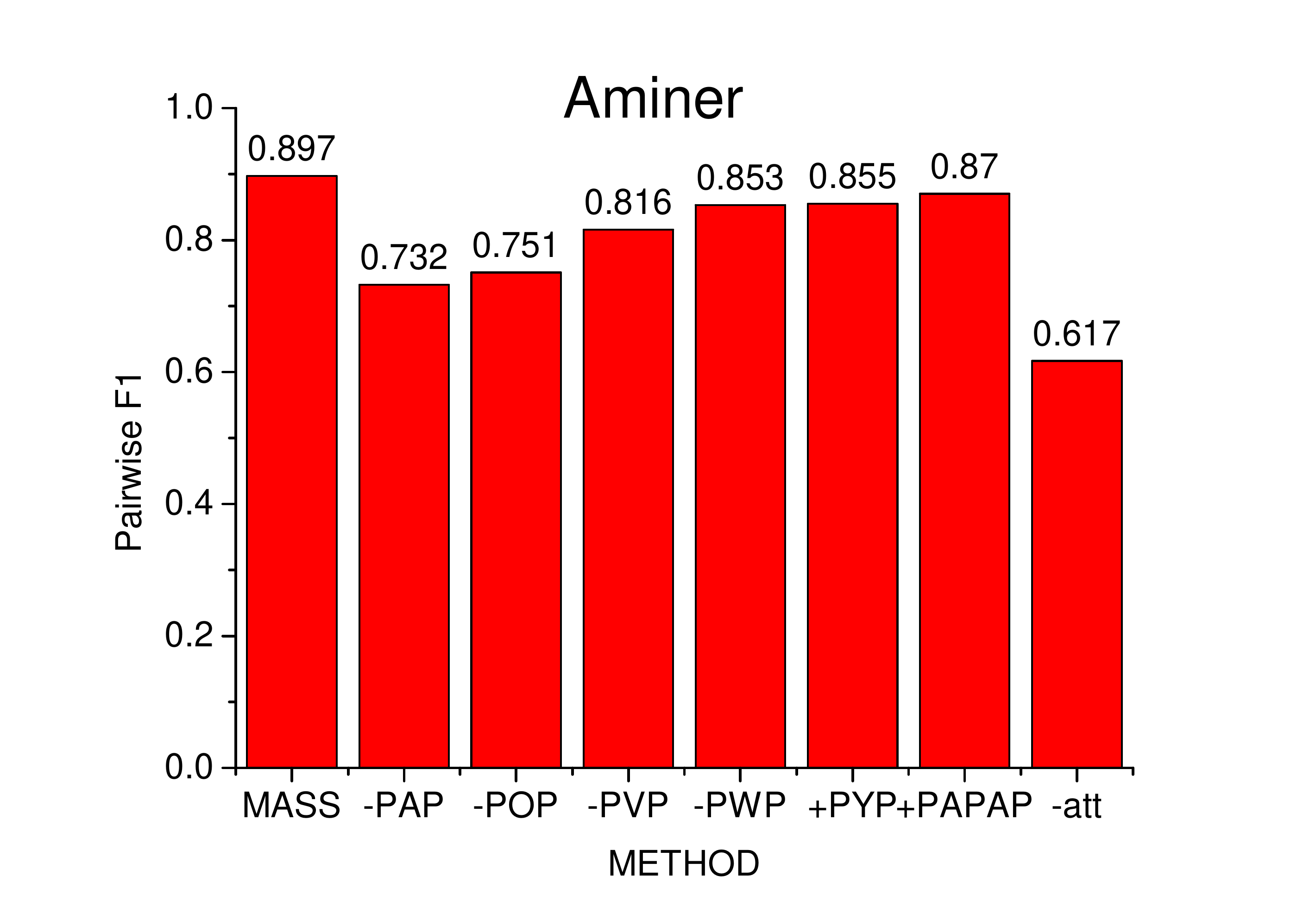}
	\caption{The performance comparison of MASS and its variants on Aminer data set. The variant method embedding without PAP are denoted as -PAP and the variant with PYP are denoted as +PYP, and so on.
\label{fig:ablation-aminer}}
\end{figure}

\begin{figure}[!htb]
	\centering
	\includegraphics[width=0.9\linewidth]{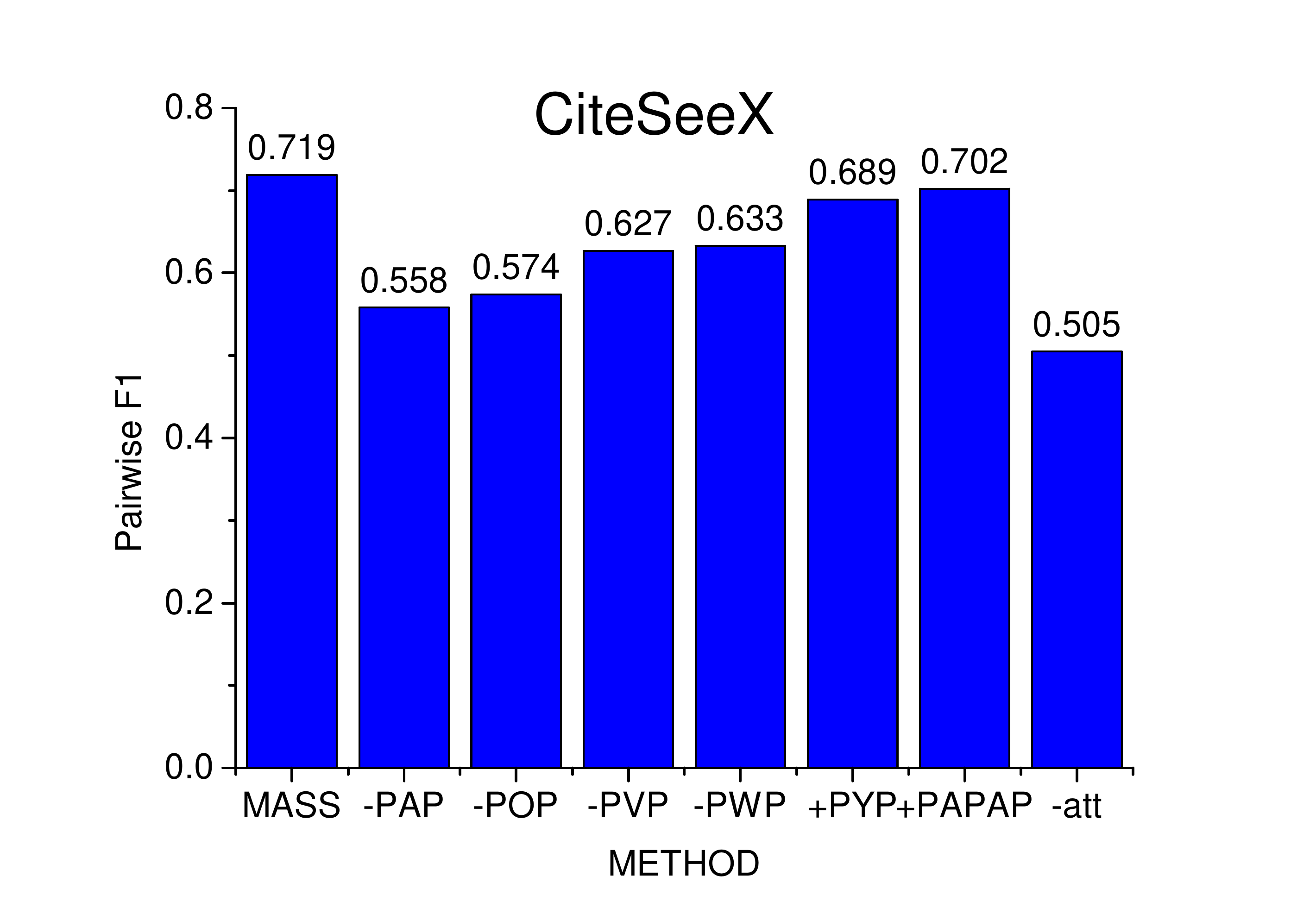}
	\caption{The performance comparison of MASS and its variants on CiteSeeX data set.
\label{fig:ablation-citeseex}}
\end{figure}

The impacts of different meta-paths on network embedding process are studied. We compare the performances of MASS and some variants which neglect one specific meta-path or import more meta-paths. To be specific, we consider the variant MASS without considering meta-path PAP, POP, PVP, PWP separately, and the variant MASS with considering PYP(Paper-Year-Paper) and PAPAP(Paper-Author-Paper-Author-Paper) which are introduced in \cite{pathsim,gui2014modeling}.
%denoted as ${\rm MASS}_{-A}, {\rm MASS}_{O}, {\rm MASS}_{-V}, {\rm MASS}_{-W}$ , and denote MASS with considering PYP(-Year-) as ${\rm MASS}_{+Y}$, PAPAP(-Author-Paper-Author-) as ${\rm MASS}_{+APA}$. 
The results are shown in Fig.\ref{fig:ablation-aminer} and \ref{fig:ablation-citeseex}. It can be found in both figures that considering more or less meta-paths will not promote the performance of MASS. The meta-path PAP provides the most impact on the F1 score compared to others, verifying the co-author relation is strong support for the two publications belonging to the same author.

The effect of importing meta-path level attention is also studied. We denote the MASS removing the attention module as $\rm -att$ in Fig.\ref{fig:ablation-aminer} and \ref{fig:ablation-citeseex}. It can be obviously found that removing the attention module makes the accuracy of the model plummet a lot. The negative impact brought by treating the features equally signifies there are different degrees of redundant information in different features and the attention technology can increase the weight of the valuable parts, making the learned features more suitable for the target task.

\section{Conclusion}
This paper proposes an effective framework for author name disambiguation task. The method uses a meta-path-based method to learn the structural representations of publications, considering the different kinds of relations between publications. Also, the meta-path level attention is employed to distinguish the importance of each relation. The publications' semantic representations are learned by the NLP method using their titles, abstracts and keywords. Then both representations are aggregated and we use an adaptive clustering method to determine the assignment of the publications. The superiority of our proposal is illustrated by the experimental results and the ablation analysis signifies the rationality of our method design. The inductive learning method can be considered in our future work to address the incremental name disambiguation task for the digital library.

\section*{Acknowledgment}
This work is supported by the Chongqing Graduate Research and Innovation Project (No.CYB22128), the Industry-University-Research Innovation Fund for Chinese Universities (No.2021ALA03016), and the National Natural Science Foundation of China (NSFC) (No.62006198). 

\bibliographystyle{IEEEtran}
\bibliography{ref}{}

\end{document}